\documentclass[10pt,twocolumn,letterpaper]{article}
\usepackage{cvpr}
\usepackage{times}
\usepackage{epsfig}
\usepackage{graphicx}
\usepackage{amsmath}
\usepackage{amssymb}
\usepackage{framed,multirow}

\usepackage[pagebackref=true,breaklinks=true,letterpaper=true,colorlinks,bookmarks=false]{hyperref}

\cvprfinalcopy 

\def\cvprPaperID{0000} 

\ifcvprfinal\fi
\begin{document}

\title{A Multi-Stream Convolutional Neural Network Framework  for Group Activity Recognition}

\author{Sina Mokhtarzadeh Azar \quad\quad Mina Ghadimi Atigh \quad\quad Ahmad Nickabadi\\
Statistical Machine Learning Lab, Amirkabir University of Technology\\
424 Hafez Ave, Tehran, Iran\\
{\tt\small \{sinamokhtarzadeh,minaghadimi,nickabadi\}@aut.ac.ir}
}

\maketitle


\begin{abstract}
In this work, we present a framework based on multi-stream convolutional neural networks (CNNs) for group activity recognition. Streams of CNNs are separately trained on different modalities and their predictions are fused at the end. Each stream has two branches to predict the group activity based on person and scene level representations. A new modality based on the human pose estimation is presented to add extra information to the model. We evaluate our method on the Volleyball and Collective Activity datasets. Experimental results show that the proposed framework is able to achieve state-of-the-art results when multiple or single  frames are given as input to the model with 90.50\% and 86.61\% accuracy on Volleyball dataset, respectively, and  87.01\% accuracy of multiple frames group activity on Collective Activity dataset.
\end{abstract}


\section{Introduction}

Human activity recognition is one of the main areas of research in computer vision where the goal is to classify what a human is doing in an input video or image. Group activity recognition is a subset of human activity recognition problem which focuses on the collective behavior of a group of people, resulted from the individual actions of the persons and their interactions. Collective activity recognition is a basic task for automatic human behavior analysis in many areas like surveillance or sports videos.

In group activity recognition, it is crucial to take into account the individual actions of the people and their interactions as in many cases the group activity is formed by these actions and interactions. So, most group activity recognition models analyzes the personal activities either explicitly or implicitly. Some works try to predict individual actions and group activity in a joint framework using probabilistic graphical models \cite{lan2012discriminative,choi2012unified} or neural networks that implement the functionality of graphical models \cite{deng2015deep,deng2016structure}. Other approaches, use pooling methods like max pooling \cite{ibrahim2016hierarchical,bagautdinov2017social} or attention pooling \cite{ramanathan2016detecting} on the individual person representations to model the relation between individual humans and the collective activity.

Another important factor in recognizing the group activity recognition is the temporal development of the individual actions and group activity. Various approaches use Recurrent Neural Networks (RNNs) to model individual actions and group activity over time \cite{ibrahim2016hierarchical,ramanathan2016detecting,tsunoda2017football,wang2017recurrent,shu2017cern,bagautdinov2017social}. This strategy provides a concrete way of modeling group activity in the video. Some existing works try to inject temporal data through Convolutional Neural Networks (CNNs) on optical flow fields calculated between two consecutive frames as an additional input to each time step of RNN \cite{wang2017recurrent,tsunoda2017football,li2017sbgar}. Recently, multi-stream convolutional networks where temporal information is modeled by CNNs on optical flow fields outperformed RNNs in action recognition task \cite{simonyan2014two,feichtenhofer2016convolutional,wang2016temporal,carreira2017quo}. To the best of our knowledge, none of the previous approaches study the group activity recognition using convolutional streams on the full range of the temporal data. When using multi-stream convolutional networks for group activity, a challenge is how to model temporal stream for multiple individuals in the input video. One can extract the sequence of images for each person according to its trajectory and compute optical flow for this sequence. Each sequence will be analyzed by a CNN on optical flow fields. This approach can be costly due to the computation of optical flow for multiple individuals and several forward passes of CNNs on the input optical flow field for all the persons in the video.

Incorporating additional information other than the appearance features has positive effect on the performance of the activity recognition models. In \cite{lan2012discriminative}, discrete pose labels are combined with action labels to form new action labels to improve accuracy. In \cite{wang2017recurrent}, a pose class is injected into the model as sub-action information.
As stated above, in \cite{wang2017recurrent,tsunoda2017football,li2017sbgar}, temporal cues obtained from the optical flow are given to the RNN. In this paper, we use the pose heatmap which contains detailed information about the body parts of the humans as novel information source in our model.


In this paper, we propose a multi-stream convolutional framework for the task of group activity recognition which addressed all of the above issues. In this framework, new input modalities are easily incorporated into the model by the addition of new convolutional streams. Optical flow, warped optical flow, pose heatmap, and the RGB frame are four different modalities considered in this paper, each of which capturing a different aspect of the input video. Moreover, to consider both the individual actions and the global activity, each stream is composed of two distinct branches; one focuses on the individual regions of the actors for individual action or group activity recognition while the other processes the whole content of the input video to recognize group activities. Evaluation on Volleyball \cite{ibrahim2016hierarchical} and Collective Activity \cite{choi2009they} datasets  shows the effectiveness of our approach.

The rest of the paper is organized as follows. First, in Sec. \ref{sec:related_works}, the related works are briefly discussed. Details of our multi-stream framework are given in Sec. \ref{sec:approach}. Sec. \ref{sec:dataset} is about the datasets. Our experimental results along with the analysis of the results are given in Sec. \ref{sec:experiments}. We conclude the paper in Sec. \ref{sec:conclusions}.

\section{Related Works}
\label{sec:related_works}
We categorize and review the previous related works in the following subsections. The first subsection present the works especially focused on group activity recognition. Since action recognition is used as a preliminary step of our group activity recognition approach, prominent works of this field are also reported in the second subsection.

\subsection{Group Activity Recognition}
There have been efforts to use probabilistic graphical models to tackle the group activity recognition problem. \cite{lan2012discriminative} propose a graphical model with person-person and group-person factors and employ a two-stage inference mechanism to find the optimal graph structure and the best possible labels for the individual actions and collective activity. In \cite{choi2012unified}, the authors explore the idea of tracking persons and predicting their activity as a group in a joint probabilistic framework. \cite{sun2016localizing} use a latent graph model for multi-target tracking, activity group localization and group activity recognition.

Initial probabilistic approaches use hand-crafted features as input to their model. With the recent success of deep neural networks in various computer vision tasks, these networks are incorporated in the probabilistic group activity recognition models as feature extractors and inference engines. \cite{deng2015deep} exploit the power of CNNs as an initial classifier to produce unary potentials. They implement the graphical model utilizing a deep neural network and perform message passing through the network to refine initial predictions. Another work explores the idea of using RNNs for message passing \cite{deng2016structure}. They also propose gating functions for learning structure of the graph.

Recently, a series of works have studied the group activity recognition using RNNs to model temporal information \cite{ibrahim2016hierarchical,ramanathan2016detecting,tsunoda2017football,wang2017recurrent,shu2017cern,bagautdinov2017social,li2017sbgar}. \cite{ibrahim2016hierarchical} employ a hierarchy of Long-Short Term Memory (LSTM) networks to predict individual actions and collective activity. In \cite{ramanathan2016detecting}, individual actors and the collective event are modeled with bidirectional and unidirectional LSTMs, respectively, and higher importance is given to key actors in an event by the means of attention pooling on persons. \cite{tsunoda2017football} introduces person-centered features as the input of a hierarchical LSTM to predict futsal activity. In a different approach, \cite{wang2017recurrent} proposes a three-level model consisting of person, group and scene representations. In the person level, each person is modeled in temporal domain with an LSTM. The final representations of these LSTMs are fed into other LSTMs at the group level in which persons are divided into spatio-temporal consistent groups. The output of the group-level LSTMs are used to generate the final scene-level prediction.

\cite{shu2017cern} propose confidence-energy recurrent network in which a novel energy layer is used instead of the common softmax layer and the uncertain predictions are avoided by the computation of p-values at the same layer. \cite{bagautdinov2017social} detect persons and classify group activity in an end to end framework. To do so, a fully convolutional network provides the initial bounding boxes for persons, which are then collectively refined in a Markov Random Field. Similar to previous works, an RNN predicts the group activity. Most recently, authors in \cite{li2017sbgar} decide upon the group activity based on the captions automatically generated for the input video as semantic information.

\subsection{Action Recognition}
In all group activity recognition models studied previously, explicit or implicit recognition of individual actions was an important part of the model. However, in some applications, the ultimate goal is to detect the individual actions of the persons not their group activity. As our model benefits from the ideas of single-person action recognition models, these models are studied here.
 \cite{simonyan2014two} propose two streams of CNNs to predict human actions in an input video. The first stream is applied to the RGB image to reason about the actions considering the appearance features while the second stream uses the optical flow fields to incorporate the temporal information into the model. The final action scores are calculated with a late fusion of the probability scores obtained from two streams. \cite{feichtenhofer2016convolutional} study fusion of two streams in earlier stages of the network. They also apply a two-stream network to multiple chunks of the video and fuse the results to further improve the model. \cite{donahue2015long} propose the Long-term Recurrent Convolutional Networks model in which the visual features extracted from the input frames by means of the CNN network are fed into an LSTM network that produces the final predictions. This model has been applied to several video processing tasks including action recognition. In a recent work \cite{li2018videolstm}, VideoLSTM is developed for action classification and localization. This model uses a 2-d attention mechanism that utilizes the motion information from optical flow fields between frames.

In \cite{wang2016temporal}, Temporal Segment Network is proposed to model long-range temporal information. Each video is divided into segments of which snippets of frames are sampled. A differentiable segmental consensus of CNNs on different segments is performed which makes end-to-end joint training of segment CNNs possible. They train their model with RGB, optical flow, and warped optical flow modalities separately and combine class scores of multiple streams at the end. The study of the effect of pretraining different architectures on large scale video classification datasets before training on smaller datasets in \cite{carreira2017quo} proves the positive impact of this step in action recognition task. Finally, it is worth mentioning that a common aspect of all above state-of-the-art action recognition methods is the use of multiple input streams.


\begin{figure*} 
\centering
\includegraphics[trim=0 130 30 0,clip,height=15cm]{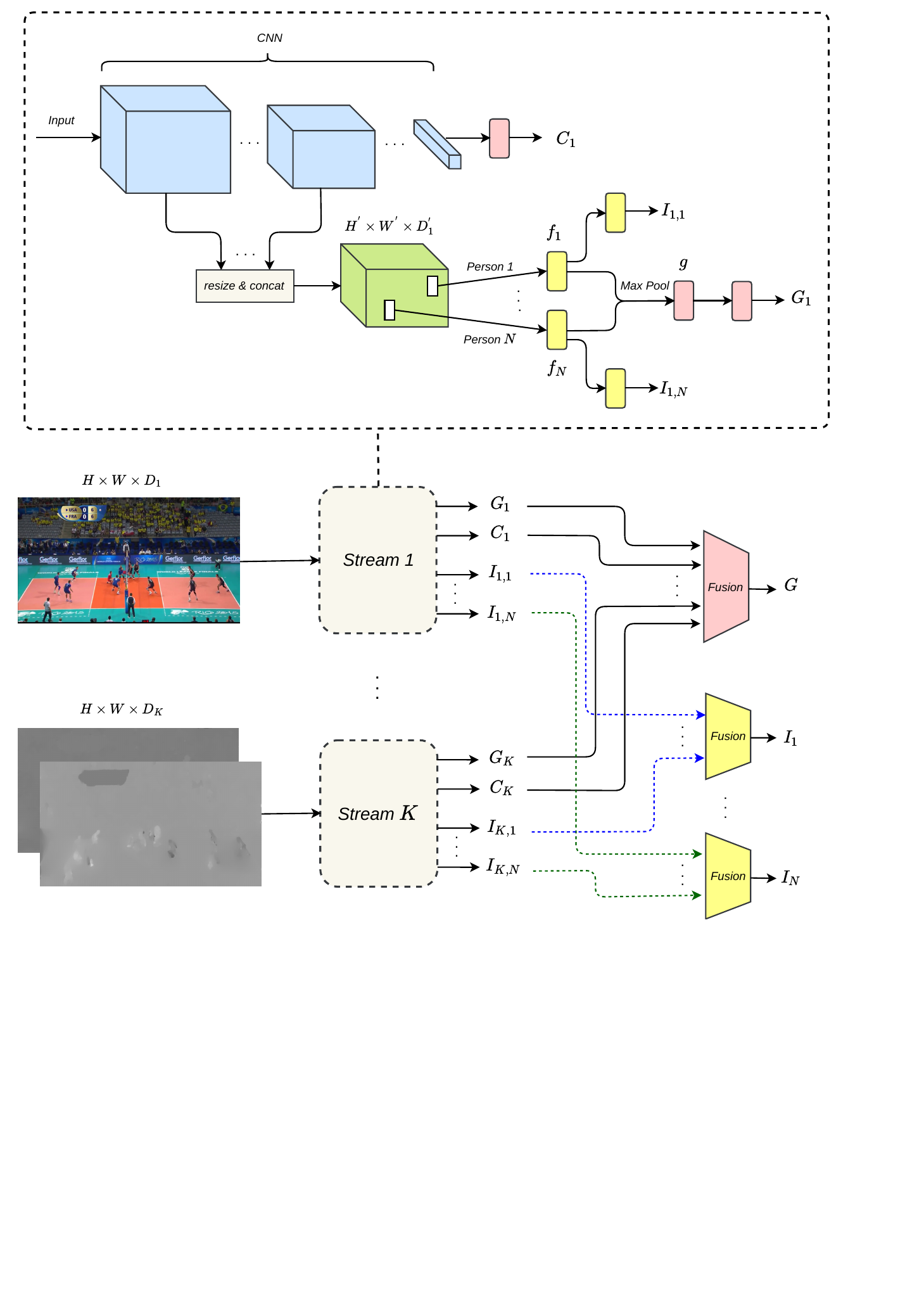}
\caption{The overall process of the proposed multi-stream convolutional framework: Each stream operates on a different modality and predictions of all streams are combined to generate the final outputs. Inside each stream, features from multiple layers of CNN are resized and concatenated together to produce the feature map from which person regions are extracted  to be used in person level group activity prediction. Scene level group activity predictions are also produced with the output of the last layer of the CNN.}
\label{fig:model}
\end{figure*}

\section{The Proposed Framework}
\label{sec:approach}

The goal of our work is to develop a framework that can easily incorporate information from new modalities as additional convolutional streams for group activity recognition. Group activity is a high level concept and existing multi-stream models for action recognition where each CNN stream reasons over the whole input without considering individuals, don't perform well. A group activity model has to consider individual actors in the process of predicting the collective behavior. Each stream in our framework learns to reason in two different ways to both make use of the person level and contextual level information. We modify the CNN of every stream to make predictions based on person representations and scene level cues simultaneously without the need to apply two separate CNNs for each part.

In the following, we first present the outline of the proposed framework and then review the modalities used in this paper.

\subsection{The Overall Architecture}
Fig.~\ref{fig:model} shows the architecture of the model. For an input video with $H \times W$ frame size and a set of $K$ input modalities, $K$ convolutional streams are considered. Each stream $k$ takes the $k$th modality with the dimensions $H\times W\times D_k$ as input to the model, where $D_k$ is the number of channels of the input, and outputs the set of predictions \{$I_{k,1}, I_{k,2},..., I_{k,N}$\} for individual actions, person level group activity $G_k$, and scene level group activity $C_k$. $I_{k,n}$ shows the prediction of stream $k$ for the action of person $n$ and $N$ is the number of persons present in the input video. These predictions need to be combined across different modalities to produce the final predictions of the multi-stream framework.
Fusion of predictions for individual actions and group activity is performed as follows:
\begin{equation}
{I}_{n} = \phi_1({I}_{1,n}, {I}_{2,n}, ... ,{I}_{K,n}) 
\end{equation}
\begin{equation}
G = \phi_2({G}_{1}, {G}_{2}, ... ,{G}_{K}, {C}_{1}, {C}_{2}, ... ,{C}_{K}) 
\end{equation}
Here, $I_n$ is the final score vector for the individual action of person $n$. ${\phi}_1$ is the fusion function to combine the set of predictions \{$I_{1,n}, I_{2,n},..., I_{K,n}$\} across different streams for person $n$. ${\phi}_2$ is the fusion function that produces final prediction vector $G$ for the group activity from the joint set $\{{G}_{1}, {G}_{2}, ... ,{G}_{K}, {C}_{1}, {C}_{2}, ... ,{C}_{K}\}$ of person pooling and scene branches of the model in different modalities. For fusion functions $\phi_1$ and $\phi_2$ we will analyze the effect of element-wise average, maximum and linear SVM.

\subsection{Convolutional Stream}
In each stream, given an input of the shape $H\times W\times D$ a CNN is first applied to produce feature maps of the form $H^{'}\times W^{'}\times D^{'}$. Like \cite{bagautdinov2017social}, feature maps from multiple layers are concatenated to form the output for the next step. According to the bounding box $B_{n}$ of each person $n$, the region corresponding to $B_n$ is extracted from the feature map and resized to a fixed shape of $M\times M\times D^{'}$. This approach is shown to be effective in reducing computational complexity of the model because it doesn't need to do the calculations of the initial layers for every crop of the image \cite{girshick15fastrcnn}.

The extracted region of the $n$th individual is given to a fully connected layer $f_n$ to form the final representation for the corresponding person $n$. These representations are fed to separate fully connected layers which classify the individuals' actions. To predict the group activity, we need to pool information from individual actors. Therefore, vector $g$ is produced using max pooling over all individual feature vectors ($f_n$s) to form a general descriptor of all persons. Another network layer is used to give the final group activity prediction of the current stream ($G_k$) based on this overall representation. It can also be helpful to let the model reason over the total spatial range of the input. This way, it can extract contextual information that wasn't available in the case of using only the bounding boxes of persons. Therefore, we also use the standard flow of layers for the chosen CNN architecture as a separate branch to directly predict the group activity. Features for individual persons are extracted from feature maps of the same CNN and we don't apply two different networks for group activity prediction from individual and scene level representations. The stream $k$ of the model is trained using $\mathcal{L}_{total}$ loss function defined as follows:
\begin{equation}
  \mathcal{L}_{total} = w_I \mathcal{L}_I + w_G \mathcal{L}_G + w_{G_C} \mathcal{L}_{G_C} 
\end{equation}
\begin{equation}
  \mathcal{L}_{I}  = -\frac{1}{N N_I}\sum_{n=1}^{N}\sum_{i=1}^{N_I} \hat{I}_{n}^{i} log\left({I}_{k,n}^{i}\right) 
\end{equation}
\begin{equation}
  \mathcal{L}_{G}  = -\frac{1}{N_G}\sum_{i=1}^{N_G} \hat{G}^{i} log\left({G}_{k}^{i}\right) 
\end{equation}
\begin{equation}
  \mathcal{L}_{G_C}  = -\frac{1}{N_G}\sum_{i=1}^{N_G} \hat{G}^{i} log\left({C}_{k}^{i}\right) 
\end{equation}
where $\mathcal{L}_{I}$, $\mathcal{L}_{G}$, and $\mathcal{L}_{G_C}$ are losses for individual actions, group activity from pooled representations, and group activity inferred from the whole image respectively. $\hat{I}_{n}^{i}$ is the ground truth value for individual action of person $n$ and action class $i$. This value is $1$ if action $i$ is the ground truth and $0$ otherwise. ${I}_{k,n}^{i}$ is the probability produced for action $i$ of $n$th person in $k$th stream. $\hat{G}^{i}$ shows the value of ground truth for group activity class $i$. ${G}_{k}^{i}$ and ${C}_{k}^{i}$ are the predicted values by the person and scene branches for group activity class $i$ of the stream $k$, respectively. The number of persons in the input video is illustrated by $N$. $N_I$ and $N_G$ show the number of classes for individual action and group activity. The total loss function is defined as a weighted sum of the above three loss functions and the weights $w_I$, $w_G$, and $w_{G_C}$ are used to control the importance of each part. 
\subsection{Input Modalities}
Every stream of the proposed framework is aimed at capturing a different aspect of the input data that would be hard to be extracted from other modalities. Separating  different modalities and assigning each modality to a different stream eases model's job to reason over specific information present in each of them. In this paper, we use two types of inputs, spatial ones that work on single frames and temporal inputs that use multiple frames.

As appearance is very important in classifying actions, a spatial stream is used to process frames in RGB format. In addition to RGB, we propose posemap as a new spatial stream, representing the locations of the body parts of actors. Body parts of a person take different structures for different actions and hence learning these relations allows the model to discriminate between actions. The posemap is represented in the form of a heatmap in which the positions of the body parts are given in a $H \times W$ map whose values are higher near the locations of the parts. As stated above, the pose of each individual person and the posemap of the whole frame are investigated in two parts of the corresponding stream.

\begin{figure*}
\centering
\includegraphics[height=6.5cm]{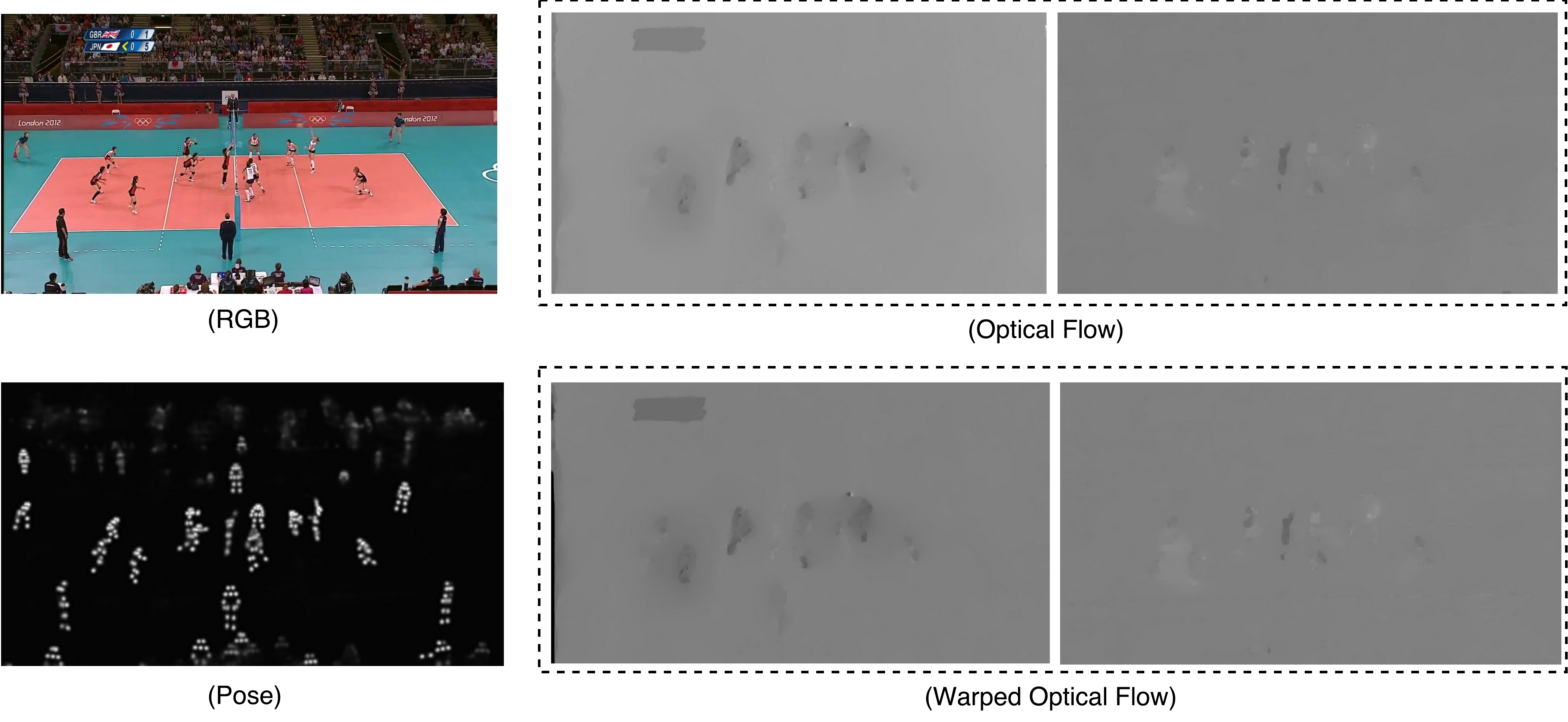}
\caption{Samples of different modalities considered in our model. We use RGB, pose, optical flow, and warped optical flow. Due to small temporal difference of adjacent frames, it is not easy to spot the differences of optical flow and warped optical flow fields in this case.}
\label{fig:modalities}
\end{figure*}

Similar to \cite{wang2016temporal}, we process optical flow and warped optical flow fields between adjacent frames in temporal streams. Optical flow is a good way of extracting motion information from a sequence of frames but it does not consider camera motion. Warped optical flow is proposed in \cite{wang2016temporal} to remove camera motion. Fig.~\ref{fig:modalities} shows samples of all the modalities used in this work. It is evident that cues about human body are extracted in posemap. We don't get ground truth heatmaps as input to our model. Posemaps have to be generated on the fly by existing models. Thanks to recent advances in human pose estimation \cite{wei2016convolutional,cao2017realtime}, we can use pretrained pose estimation models to extract posemap from an input image.

\section{Datasets}
\label{sec:dataset}
We evaluate our model on Volleyball \cite{ibrahim2016hierarchical} and Collective Activity  \cite{choi2009they} datasets.
\subsection{Volleyball Dataset}
\begin{figure}
\centering
\includegraphics[trim=120 30 100 30,clip,height=5.6cm]{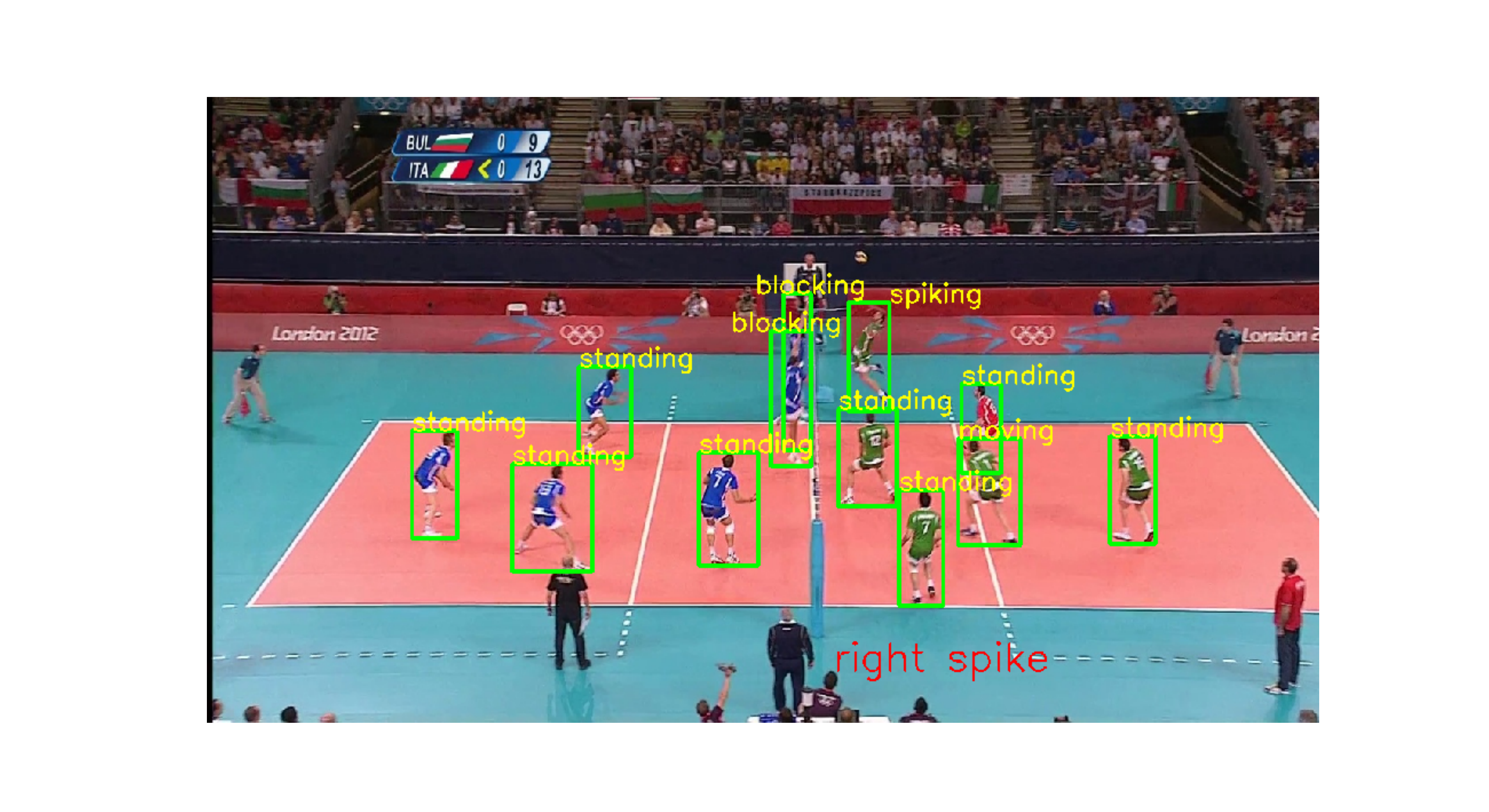}
\caption{Example middle frame of a clip from Volleyball Dataset \cite{ibrahim2016hierarchical}. Bounding boxes, actions of persons, and the group activity are specified on the image.}
\label{fig:sample}
\end{figure}

 This dataset consists of 55 videos of volleyball matches. Every match is broken into multiple smaller length video clips of specific group activities. Videos of 39 matches are used for training and the other 16 matches are used for testing. This means we have 3493 clips for training and 1337 for testing the model. Each clip has 41 frames with middle frame labeled with group activity, actions of persons and bounding box coordinates for each person. Other frames don't have any labels. In previous works \cite{ibrahim2016hierarchical,wang2017recurrent,shu2017cern} a tracker is used to find bounding boxes of persons in other frames but our approach doesn't need to know the location of individuals in frames other than the middle one. There are 8 group activity and 9 individual action classes in this dataset. Group activity labels are \textit{right set}, \textit{right spike}, \textit{right pass}, \textit{right winpoint}, \textit{left set}, \textit{left spike}, \textit{left pass}, and \textit{left winpoint}. Individual actions are \textit{blocking}, \textit{digging}, \textit{falling}, \textit{jumping}, \textit{moving}, \textit{setting}, \textit{spiking}, \textit{standing}, and \textit{waiting}. A sample frame of the dataset with bounding boxes along with individual actions and group activity labels are shown in Fig.~\ref{fig:sample}.

\subsection{Collective Activity Dataset}
In this dataset, there are 44 videos with different number of frames in each video. 31 videos are used for training and 13 for testing. Person bounding boxes and actions are labeled every 10th frame. Therefore, the train and test sets are consisted of 1908 and 639 short clips, respectively. Action labels are \textit{crossing}, \textit{waiting}, \textit{queuing}, \textit{walking}, and \textit{talking}. The most frequent person action label in each short clip is considered as the group activity label for that clip.

\section{Experiments}
\label{sec:experiments}
\subsection{Implementation Details}
We implement our model with Tensorflow \cite{abadi2016tensorflow}. We use Inception-V3 \cite{szegedy2016rethinking} as our CNN architecture. Inception-V3 has 3 different types of inception modules and each of them is repeated for a number of times (3, 5, and 2 times). There are a total of 10 inception modules. Final output of two initial inception types are resized and concatenated to form a $H^{'} \times W^{'} \times D^{'}$ feature map. Final output of type-1 inception module has a depth of 256. This depth is 768 for the output of type-2 inception. Therefore, the depth of featuremap will be 1056. 

\textbf{Volleyball Dataset}. Input frames are resized to $720\times1280$. Then, the dimensions of the featuremap will be $87\times157\times1056$. The cropped region from the featuremap for each person is resized to $4\times4$. The number of neurons in $f_i$ is set to 2048. We observe that values of 2, 1, and 1 for weights $w_I$, $w_G$, and $w_{G_C}$ produce consistent results. Adam optimizer \cite{kingma2014adam} with a learning rate of 0.00001 is used to train all models. Data augmentation is not performed.

\textbf{Collective Activity Dataset}. Resized images have the shape $480\times720$ and their featuremap dimensions are $57\times87\times1056$. For optical flow, warped optical flow, rgb, and posemap streams, we use $f_i$ with 512, 512, 512, and 256 neurons, respectively. $w_I$, $w_G$, and $w_{G_C}$ are all set to 1. The learning rate for Adam optimizer is set to 0.00001 for optical flow and warped optical flow streams and 0.0001 for rgb and posemap streams. Due to small number of data samples and the large variations in background, we do random horizontal flips as data augmentation.

To compute optical flow and warped optical flow we resort to the same approach presented in \cite{wang2016temporal} and use TVL1 algorithm \cite{zach2007duality}. In addition to the middle frame, we take 4 frames before and 5 frames after it to form the temporal domain. Both temporal modalities are extracted prior to train and compressed and saved as JPEG files to speed up the training. To extract pose images, we use pretrained model of \cite{cao2017realtime} on COCO 2016 keypoints challenge dataset \cite{lin2014microsoft}. We use only a scale of 2 and construct the pose image using background heatmap in the model's output. Background has higher values for the points where there is no human body part. We subtract every point in the heatmap from maximum value of the heatmap and use them as new values for heatmap to give higher values to points near human body parts. These points are scaled to the range of [0, 255] and saved like flow fields.

We initialize all layers of Inception-V3 with pretrained weights on ImageNet. In temporal streams input has more than three channels. To initialize inputs of each channel, we follow the same approach proposed in \cite{wang2016temporal} and initialize every channel in temporal streams with the channel-wise mean of pretrained Inception-V3. Posemaps have only one channel and we repeat that channel to construct an input with three channels to make it possible to use pretrained weights on Imagenet.

\subsection{Model Performance Analysis}
In this subsection, different strategies that lead to the best performing model on Volleyball dataset are analyzed. As we discussed previously, using person level representations to predict group activity discards the contextual information. We expect adding scene level predictions to previous predictions to improve the results. Table.~\ref{pool-scene} shows the experimental results of adding scene predictions for each stream. Although accuracies of scene level predictions are lower than predictions in person level branch, the positive effect of additional contextual cues is evident in all modalities. Accuracy of the scene branch in pose modality is significantly lower than performance of different branches in other modalities. Although it leads to a minor increase in accuracy of the fused pose stream, it can cause performance degradation in combination with other modalities. We will discuss this issue when we study the impact of the pose modality on the performance of the model.

\begin{table}[]
\begin{center}

\begin{tabular}{|l|c|c|c|}

\hline
{Modality} & Person & Scene & Avg \\ \hline
RGB                            & 84.74      & 82.72       & 85.41      \\ \hline
Optical Flow                   & 85.26      & 83.84       & 85.93      \\ \hline
Warped Optical Flow            & 84.89      & 83.99       & 86.91      \\ \hline
Pose                           & 82.27      & 76.43       & 82.64      \\ \hline
\end{tabular}
\end{center}
\caption{Accuracy of different modalities with group activity predictions produced from person level representations along with scene level predictions and their fusion by average on Volleyball dataset.}
\label{pool-scene}
\end{table}

As we mentioned earlier, in this framework new convolutional streams on new modalities can be integrated into final model to improve the performance. To evaluate the effect of additional modalities, we experiment with different combinations of streams. The results are summarized in Table.~\ref{combinations}. Starting with RGB modality we fuse its predictions with optical flow stream's output and the improvement over spatial RGB stream is approximately 4\% that shows the power of optical flow fields in extracting temporal cues from a sequence of frames. Warped optical flow stream works on a similar input to optical flow stream and adding it to model only dissolves minor shortcomings of RGB and optical flow fusion. 

Previously, we discussed the positive effect of contextual information in the scene branches of the model. Here, we add them to the multi-stream framework. Increase of accuracy becomes bold again after adding the RGB scene branch predictions. Temporal streams have minor improvement relative to first scene branch. Intuitively, first scene stream encodes contextual information that is highly similar to other scene streams. Therefore, adding new scene streams improves the accuracy, but it is not noticeable relative to the total accuracy.

Next, we consider evaluating the effect of our new modality, posemap. In Table.~\ref{pool-scene} accuracies of both person and scene branches of posemap stream are reported. Person based branch is able to produce an acceptable result but the scene branch lacks the power of extracting contextual information in this case. A downgrade in performance is expected for the scene predictions because in the posemaps generated from the Volleyball Dataset, information about some of the objects in the context including ball, net, and court are omitted. One of the flexibilities of our framework is that depending on the modality, we can choose to use both or one of the person or scene branches. For example, in Table.~\ref{combinations}, we observe that adding group activity predictions produced by person level branch of the posemap to other modalities improves the accuracy but including the scene branch predictions decreases the accuracy due to the poor performance of scene predictions in this specific modality. Therefore, we choose not to use scene branch predictions of posemap in our final model for Volleyball dataset.

\begin{table}[]
\begin{center}
\begin{tabular}{|l|c|}
\hline
\hspace{0.1cm}Combination\hspace{0.1cm}                   & \multicolumn{1}{l|}{\hspace{0.1cm}Accuracy\hspace{0.1cm}} \\ \hline
\hspace{0.1cm}RGB (Person)\hspace{0.1cm}                    & 84.74                         \\ \hline
\hspace{0.1cm}+ Optical Flow (Person)\hspace{0.1cm}         & 88.55                         \\ \hline
\hspace{0.1cm}+ Warped Optical Flow (Person)\hspace{0.1cm}  & 88.85                         \\ \hline
\hspace{0.1cm}+ RGB (Scene)\hspace{0.1cm}                 & 89.60                          \\ \hline
\hspace{0.1cm}+ Optical Flow (Scene)\hspace{0.1cm}        & 89.75                         \\ \hline
\hspace{0.1cm}+ Warped Optical Flow (Scene)\hspace{0.1cm} & 89.82                         \\ \hline
\hspace{0.1cm}+ Pose (Person)\hspace{0.1cm}                 & \textbf{90.42}                \\ \hline
\hspace{0.1cm}+ Pose (Scene)\hspace{0.1cm}                & 89.90                         \\ \hline
\end{tabular}
\end{center}
\caption{Exploration of the impact of different modalities on the accuracy of the model on Volleyball dataset. Predictions of the branches of each modality are added from top to bottom.}
\label{combinations}
\end{table}

To fuse predictions of different modalities, we can take element-wise maximum or average of predictions made by scene and person branches of all streams. We also experiment with fusion using a linear SVM. The results for individual actions and group activity are summarized in Table.~\ref{fusion}. We analyze different fusion methods for our final model along with the fusion approaches for single frame model where we generate all optical flow and warped optical flow related predictions. Fusion by maximum has poor performance in all cases. Average fusion has better results compared to maximum. Weighted sum using the linear SVM has the best results in every part. Due to the consistency of SVM, we choose it as fusion method in our final method.

\begin{table}[]
\begin{center}

\begin{tabular}{|c|c|c|c|c|}
\hline
\multirow{2}{*}{Method} & \multicolumn{2}{c|}{Multiple Frames} & \multicolumn{2}{c|}{Single Frame} \\ \cline{2-5} 
                        & Individual         & Group         & Individual         & Group        \\ \hline
\multicolumn{1}{|l|}{Max}                    & 82.65               & 88.85         & 81.04              & 85.71        \\ \hline
\multicolumn{1}{|l|}{Avg}                      & 83.06               & 90.42         & 81.32              & 86.38        \\ \hline
\multicolumn{1}{|l|}{SVM}                      & \textbf{83.15}               & \textbf{90.50}          & \textbf{81.56}              & \textbf{86.61}        \\ \hline
\end{tabular}
\end{center}
\caption{Comparison of different fusion methods on Volleyball dataset. Two cases where model can use multiple frames and where it gets a single frame are shown separately. Both individual actions and group activity are reported for each case.}
\label{fusion}
\end{table}

\begin{table*}[]
\begin{center}

\begin{tabular}{|l|c|c|c|c|}
\hline
\multicolumn{1}{|c|}{\multirow{2}{*}{Method}} & \multicolumn{2}{c|}{Multiple Frames} & \multicolumn{2}{c|}{Single Frame} \\ \cline{2-5} 
\multicolumn{1}{|c|}{}                        & Individual        & Group          & Individual         & Group        \\ \hline
HDTM \cite{ibrahim2016hierarchical}                                         & -                   & 81.90           & -                  & -            \\ \hline
CERN \cite{shu2017cern}                                         & 69.10                & 83.30           & -                  & -            \\ \hline
Social Scene \cite{bagautdinov2017social}                                & 82.40                & 89.90           & 81.10               & 83.80         \\ \hline
Ours                                         & \textbf{83.15}               & \textbf{90.50}           & \textbf{81.56}              & \textbf{86.61}      \\ \hline
\end{tabular}
\end{center}
\caption{Comparison of our results with those of the state-of-the-art methods in cases where models have access to multiple frames or only a single frame.}
\label{table:state_of_the_art}
\end{table*}

\subsection{Comparison with State-of-the-Art}

\subsubsection{Volleyball Dataset}
As we discussed in the previous section, using both person and scene level predictions in RGB, optical flow, and warped optical flow combined with person level predictions of posemap is able to achieve highest performance and we choose this combination as our final model. Our results along with the results of prior works are summarized in Table.~\ref{table:state_of_the_art}. We compare our model with hierarchical deep temporal model (HDTM) \cite{ibrahim2016hierarchical}, confidence energy recurrent network (CERN) \cite{shu2017cern}, and social scene understanding (Social Scene) \cite{bagautdinov2017social}. Average accuracy of individuals and group activity in two cases where single or multiple frames are fed to the model are reported. Our model along with Social Scene model achieve significantly better accuracies for group activity when a sequence of frames is used. Our approach slightly outperforms Social Scene in this case. Same is true for the accuracy of individual actions where our model performs 0.7 better than Social Scene.

These observations explain three points. First, temporal streams are able to model temporal information at least as good as the recurrent models. Second, even without having access to the trajectory of persons, model can learn to extract appropriate features from flow fields with regard to the position of persons in the middle frame and achieve competitive results with state-of-the-art model. Third, posemap includes valuable information that is not captured by other streams. This helps the model to perform better than the state of the art. Furthermore, by analyzing results on single frame we observe a significant improvement over Social Scene in both individual and group activity in terms of accuracy. These observations further approve the effectiveness of the pose modality and our model's ability to learn from it.

\subsubsection{Collective Activity Dataset}
For Collective Activity dataset, we use all four streams with their person-level and scene-level predictions and combine them with a linear SVM.
Group activity performance of the proposed model is then evaluated in two different settings. 
In the first experiment, the Multi-class Classification Accuracy (MCA) of the proposed model over the five classes of Collective Activity dataset is compared with those of seven other group activity recognition methods (Table.~\ref{table:collective}).
As the reported results of this table show, the proposed model has competitive results with the best performing method \cite{shu2017cern} and outperforms other approaches. However, a small modification of the dataset clearly depicts the superiority of the proposed model.
\textit{Crossing} activity is basically a form of \textit{Walking} activity performed by a group of persons walking across a street. Therefore, similar to \cite{wang2017recurrent}, we merge the predictions of \textit{Crossing} and \textit{Walking} activities into the new \textit{Moving} activity and report the mean per class accuracy (MPCA) in Table.~\ref{table:collective_moving}. We use the reported accuracies in \cite{wang2017recurrent} for \cite{choi2012unified}, \cite{hajimirsadeghi2015visual}, \cite{lan2012discriminative}, and \cite{ibrahim2016hierarchical}. The MPCA for \cite{li2017sbgar} is calculated from their reported confusion matrix. We were not able to report the results for \cite{shu2017cern} due to the lack of the confusion matrix in this work. As the results show, our model is significantly better than other methods in this setting and it is evident that most of its mistakes occur when it tries to discriminate between \textit{Crossing} and \textit{Walking} activities and after combining them into the \textit{Moving} activity, high performance is observed for both \textit{Moving} activity classification accuracy and MPCA of all classes.

\begin{table}[]
\begin{center}
\begin{tabular}{|l|c|}
\hline
\hspace{0.1cm}Method\hspace{0.1cm}                   & \multicolumn{1}{l|}{\hspace{0.1cm}Accuracy\hspace{0.1cm}} \\ \hline
\hspace{0.1cm}\cite{choi2012unified}\hspace{0.1cm}         & 80.40                         \\ \hline
\hspace{0.1cm}\cite{hajimirsadeghi2015visual}\hspace{0.1cm}  & 83.40                         \\ \hline
\hspace{0.1cm}\cite{lan2012discriminative}\hspace{0.1cm}                 & 79.70                        \\ \hline
\hspace{0.1cm}\cite{ibrahim2016hierarchical}\hspace{0.1cm}        & 81.50                         \\ \hline
\hspace{0.1cm}\cite{li2017sbgar}\hspace{0.1cm} & 86.10                       \\ \hline
\hspace{0.1cm}\cite{shu2017cern}\hspace{0.1cm}                 & \textbf{87.20}                \\ \hline
\hspace{0.1cm}Ours\hspace{0.1cm}                & 87.01                         \\ \hline
\end{tabular}
\end{center}
\caption{Comparison of the MCA of our multi-stream model with the state-of-the-art approaches on Collective Activity dataset.}
\label{table:collective}
\end{table}

\begin{table}[]
\begin{center}
\begin{tabular}{|l|c|c|c|c|c|}
\hline
                  & {Moving}& {Waiting}& {Queuing}& {Talking} & {MPCA}\\ \hline
\cite{choi2012unified}         & 90.00 & 82.90 & 95.40 & 94.90  & 90.80   \\ \hline
\cite{hajimirsadeghi2015visual}& 87.00 & 75.00 & 92.00 & 99.00  & 88.30   \\ \hline
\cite{lan2012discriminative}   & 92.00 & 69.00 & 76.00 & 99.00  & 84.00   \\ \hline
\cite{ibrahim2016hierarchical} & 95.90 & 66.40 & 96.80 & \textbf{99.50}  & 89.70   \\ \hline
\cite{li2017sbgar}             & 90.77 & 81.37 & 99.16 & 84.62  & 88.98   \\ \hline
\cite{wang2017recurrent}       & 94.40 & 63.60 & \textbf{100.00} & \textbf{99.50}  & 89.40   \\ \hline
Ours                            & \textbf{96.73} & \textbf{87.67} & 97.84 & 98.36  & \textbf{95.26}  \\ \hline
\end{tabular}
\end{center}
\caption{The MPCA and per class accuracies of our model compared with other methods on Collective Activity dataset.}
\label{table:collective_moving}
\end{table}

\section{Conclusions}
\label{sec:conclusions}
In this paper, we presented multi-stream convolutional networks as a framework for group activity recognition. In this framework, new modalities both in spatial and temporal domains can be easily plugged in to improve the model's power. Depending on the input modality, one or both person and scene level group activity predictions can be used in the model. Experimental results on Volleyball and Collective Activity datasets show the effectiveness of our approach. Our model is able to surpass state-of-the-art methods in group activity recognition in both multiple and single frame cases on Volleyball dataset and shows competitive group activity recognition results on Collective Activity dataset in different settings.

{\small
\bibliographystyle{ieee}
\bibliography{egbib}
}

\end{document}